%% file: iclr2026_conference.tex
\documentclass{article} 
\usepackage{iclr2026_conference,times}

\input{math_commands.tex}

\usepackage[utf8]{inputenc} 
\usepackage[T1]{fontenc}    
\usepackage{hyperref}       
\usepackage{url}            
\usepackage{booktabs}       
\usepackage{amsfonts}       
\usepackage{nicefrac}       
\usepackage{microtype}      
\usepackage{xcolor}         

\usepackage{xspace}
\usepackage{amsmath}
\usepackage{multirow}
\usepackage{caption}
\usepackage{pifont}
\usepackage{graphicx}
\usepackage{tabularx}
\usepackage{wrapfig,lipsum}
\usepackage{subfig}
\usepackage[inline]{enumitem}
\usepackage{algorithm}
\usepackage{algpseudocode}

\usepackage[nameinlink]{cleveref}
\crefname{paragraph}{section}{sections}
\Crefname{paragraph}{Section}{Sections}
\crefname{paragraph}{Sec.}{Secs.}
\crefname{section}{Sec.}{Secs.}
\crefname{table}{Tab.}{Tabs.}
\crefname{equation}{Eq.}{Eqs.}
\crefname{figure}{Figure}{Figures}
\crefname{appendix}{Appendix}{Appendix}
\crefname{algorithm}{Algorithm}{Algorithms}
\input{preamble}

\usepackage{url}
\usepackage{hyperref}

\title{Taming Diffusion Transformer for Efficient Mobile Video Generation in Seconds}

\author{%
  Yushu Wu\textsuperscript{1,2}\thanks{Work done during an internship at Snap Inc.}\hspace{0.4em}\thanks{Equal contributions}\quad
  Yanyu Li\textsuperscript{1}\samethanks[2]\quad
  Anil Kag\textsuperscript{1}\quad 
  Ivan Skorokhodov\textsuperscript{1}\quad 
  \textbf{Willi Menapace}\textsuperscript{1}\quad \\
  \textbf{Ke Ma}\textsuperscript{1}\quad
  \textbf{Arpit Sahni}\textsuperscript{1}\quad
  \textbf{Ju Hu}\textsuperscript{1}\quad 
  \textbf{Aliaksandr Siarohin}\textsuperscript{1}\quad 
  \textbf{Dhritiman Sagar}\textsuperscript{1}\quad \\
  \textbf{Yanzhi Wang}\textsuperscript{2}\quad
  \textbf{Sergey Tulyakov}\textsuperscript{1}\\
  \textsuperscript{1}Snap Inc.\quad
  \textsuperscript{2}Northeastern University \\
  {Project Page: \href{https://snap-research.github.io/mobile_video_dit/}
  {https://snap-research.github.io/mobile\_video\_dit/}
  }
}

\preprintcopy
\begin{document}

\maketitle

\begin{figure*}[h]
  \vspace{-3em}
  \centering
  \includegraphics[width=\linewidth]{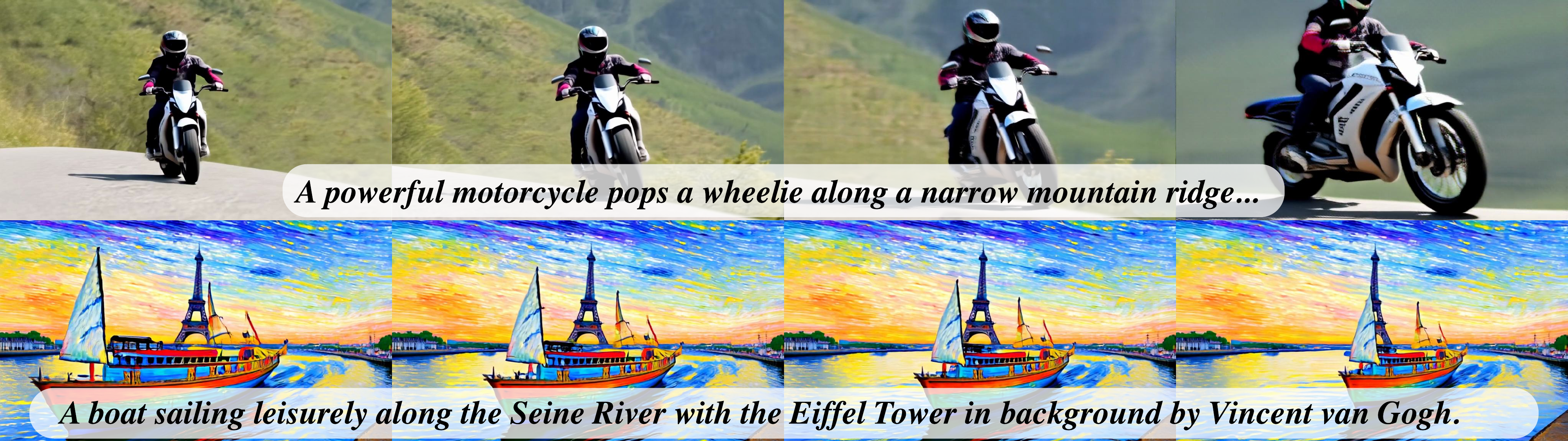}
\caption{Videos generated by our efficient Diffusion Transformer.
}
\label{fig:teaser}
\end{figure*}

\begin{abstract}

\input{tex/0_abstract}
\end{abstract}

\input{tex/1_introduction}
\input{tex/2_preliminary}

\input{tex/3_method}
\input{tex/4_evaluation}

\input{tex/5_conclusion}



\bibliography{iclr2026_conference}
\bibliographystyle{iclr2026_conference}

\newpage
\appendix
\input{tex/appendix}

\end{document}

%% file: math_commands.tex
\usepackage{amsmath,amsfonts,bm}

\def\eqref#1{equation~\ref{#1}}

\def\1{\bm{1}}

\DeclareMathAlphabet{\mathsfit}{\encodingdefault}{\sfdefault}{m}{sl}
\SetMathAlphabet{\mathsfit}{bold}{\encodingdefault}{\sfdefault}{bx}{n}



%% file: preamble.tex
\newcommand{\onedot}{\ifx\@let@token.\else.\null\fi\xspace}

\newcommand{\eg}[0]{\emph{e.g}\onedot} 
\newcommand{\ie}[0]{\emph{i.e}\onedot}

\newcommand*\samethanks[1][\value{footnote}]{\footnotemark[#1]}

%% file: tex/0_abstract.tex
Diffusion Transformers (DiT) have shown strong performance in video generation tasks, but their high computational cost makes them impractical for resource-constrained devices like smartphones, and practical on-device generation is even more challenging. 
In this work, we propose a series of novel optimizations to significantly accelerate video generation and enable practical deployment on mobile platforms.
First, we employ a highly compressed variational autoencoder (VAE) to reduce the dimensionality of the input data without sacrificing visual quality. 
Second, we introduce a KD-guided, sensitivity-aware tri-level pruning strategy to shrink the model size to suit mobile platforms while preserving critical performance characteristics.
Third, we develop an adversarial step distillation technique tailored for DiT, which allows us to reduce the number of inference steps to four.
Combined, these optimizations enable our model to achieve approximately 15 frames per second (FPS) generation speed on an iPhone 16 Pro Max, demonstrating the feasibility of efficient, high-quality video generation on mobile devices.

%% file: tex/1_introduction.tex
\section{Introduction}

The rapid advancement of generative models~\citep{RecFlow,stablediffusion3,LDM} has led to significant breakthroughs in video generation~\citep{SVD,sora,opensora,wan2025wanopenadvancedlargescale,CogVideoX,genmo2024mochi,HunyuanVideo}, with Diffusion Transformers emerging as one of the most effective architectures for producing temporally coherent and visually compelling video content.
These models leverage the strengths of diffusion processes~\citep{LDM,DDPM,RecFlow} for stepwise refinement and transformer-based attention for capturing long-range dependencies across frames, making them particularly suitable for generating complex, high-fidelity video sequences.
As such, they have become a cornerstone in state-of-the-art video synthesis pipelines.

Despite their impressive generative capabilities, Diffusion Transformers suffer from substantial computational overhead, especially when applied to high-resolution video generation. 
While this brings significant quality improvements, the computation and memory consumption of the 3D full attention \citep{CogVideoX,HunyuanVideo} scale quadratically with respect to total tokens ($t\times H \times W$).
This limitation poses a critical challenge for deploying these models in interactive settings, particularly on mobile devices with limited processing power and energy budgets.
Existing efforts to optimize diffusion-based models, such as step reduction~\citep{zhang2024sfv,li2024snapfusion,dmd2}, efficient backbone \citep{wu2024snapgenvgeneratingfivesecondvideo,yahia2024mobilevideodiffusion,zhao2024mobilediffusion}, are mainly focused on UNet-based denoisers, which are naturally less expressive. Very few work~\citep{LTX-video} investigates the efficiency of DiT, and often suffers from perceptual quality or temporal consistency loss.
Furthermore, most current acceleration methods are designed for desktop~\citep{LTX-video,wan2025wanopenadvancedlargescale}, and do not translate well to edge devices.

\input{table/wrap_table_teaser}
In this work, we present a comprehensive optimization pipeline tailored specifically to accelerate video diffusion transformers for mobile deployment.
Our approach combines four key strategies. 

\textbf{(A) High-Compression Video Variational Autoencoder (VAE).} 
First, we investigate the compression rate of the VAE.
High compression video VAE can significantly reduce the number of tokens in the latent representation, thus speeding up DiT inference.
However, VAEs with an aggressive compression ratio often suffer from a loss of reconstruction quality, which likely leads to a loss in diffusion model generation quality. 
The trade-off between compression ratio and diffusion quality remains underexplored for on-device video generation. 
In this work, we create a series of video VAEs with different compression ratios and compare the speed gain versus generation quality loss. 
We have several findings:
(i) The reconstruction and diffusion generation quality corresponds well with the compression ratio.
(ii) The speedup from the higher VAE compression ratio is significant. 
(iii) Though slightly degraded, we can still find a sweet point that balances speed and quality.

\textbf{(B) Efficient Mobile DiT.}
Second, we find that directly training a lightweight DiT designated for mobile is challenging.
Instead, we start from a larger pre-trained supernet and propose a sensitivity-aware tri-level pruning with a KD-Guided framework that selectively removes less critical components of the model based on their contribution to both runtime and output quality. 
This pruning reduces the number of DiT blocks, feed-forward features, and attention heads. 
The final architecture has $915$M parameters and can be easily deployed on a modern device such as an iPhone 16 Pro Max.
Further, we improve the pruned model performance by aligning features of the pruned network and the supernet through knowledge distillation.

\textbf{(C) Adversarial Step Distillation.} 
Third, we design a new discriminator head tailored for adversarial step-distillation on DiTs that achieves full-step quality with only a few sampling steps.
Prior adversarial distillation methods for video diffusion models mainly focus on UNet backbones~\citep{yahia2024mobilevideodiffusion,zhang2024sfv} and less challenging image-to-video tasks~\citep{SVD} and do not transfer directly to DiTs.
Our discriminator design inherits the first $K$ frozen pretrained generator blocks as a time-conditioned feature parser, and adds learnable with self-attention and cross-attention to fully capture conditions. 
This design enables \emph{four-step} generation without classifier-free guidance (CFG), yielding $20\times$ faster inference than a typical 40-step CFG recipe.

\textbf{(D) Operator Optimization for Efficient Inference.}
Finally, we identify a memory bottleneck in the linear layer of DiTs~(\ie feed-forward network), where the limited bandwidth prevents operators from approaching their theoretical speed on device.
To address this, we introduce a tiled GEMM strategy that alleviates the memory bottleneck without requiring kernel-level modifications.
The design achieves over 50\% speedup on targeted linear layers and $\sim$10\% acceleration for DiT inference.

With these optimizations, our model can generate high-quality video at over $15$ frames per second (FPS) on an iPhone 16 Pro Max using only four denoising steps.
Extensive experiments demonstrate that our method maintains strong visual fidelity and temporal consistency, closely matching the outputs of full-resolution, unpruned models.
For the first time, our work advances the state-of-the-art for on-device efficient video generation by making practical diffusion-based video synthesis feasible on consumer-grade mobile hardware. Our contributions can be summarized as follows, 
\begin{itemize}[leftmargin=1em,itemsep=0pt]
    \item We are the first to systematically investigate the trade-off between latent compression ratio, generation quality, and speed for on-device video generation. We find that for diffusion transformer, a $8\times 32\times 32$ VAE achieves a good trade-off between generation speed and quality.  
    
    \item To obtain an efficient DiT backbone, training a smaller network from scratch gives inferior results. Instead, we start from a large pre-trained super network and apply distillation-guided, sensitivity-aware pruning, yielding a compact network with optimized depth and width.
    
    \item For adversarial step-distillation, we propose a new discriminator design tailored for DiTs which outperforms prior methods by a large margin. We achieve $4$-step inference without CFG. 

    \item We identify the memory bottleneck in the feed-forward layers of DiTs on device performance and introduce a tiled GEMM strategy that alleviates this issue, enabling more efficient and hardware friendly on-device inference.
\end{itemize}

%% file: table/wrap_table_teaser.tex
\newlength{\oldintextsep}
\setlength{\oldintextsep}{\intextsep}
\setlength{\intextsep}{0.5em} 

\begin{wraptable}{r}{0.45\linewidth}
\vspace{-1.2em}
\small
\centering
\captionsetup{aboveskip=2pt,skip=2pt} 
\caption{Our model is the first DiT-based mobile video generator. Generation speed is reported as FPS. See~\cref{sec:experiments} for details. }
\label{tab:speed}
\resizebox{\linewidth}{!}{
\begin{tabular}{ccccc}
\toprule
Model       & Params (B) & VBench & A100 & iPhone \\
\midrule
Wan2.1      & 1.3        & 83.33  & 0.2  & \ding{55}      \\
LTX         & 1.8        & 80.00  & 6.1  & \ding{55}      \\
\hline
Ours-Server & 2.0        & 83.09  & 6.4  & \ding{55}      \\
Ours-Mobile & 0.9        & 81.45  & 151.3  & $\sim$15   \\
\bottomrule
\end{tabular}
}
\end{wraptable} 

\setlength{\intextsep}{\oldintextsep}

%% file: tex/2_preliminary.tex
\section{Related Work}
\textbf{Video Diffusion Models.} 
Recent years have seen rapid progress in video generation models~\citep{wan2025wanopenadvancedlargescale,sora,opensoraplan,CogVideoX,HunyuanVideo,genmo2024mochi,kling,StepVideoT2V}. 
Most advances focus on large diffusion models that iteratively denoise Gaussian noise into realistic videos, conditioned on text or images. 
These approaches include pixel-space models~\citep{SnapVideo,ImagenVideo} and latent-space models~\citep{wan2025wanopenadvancedlargescale,CogVideoX}. 
While such systems~\citep{sora,opensora,SnapVideo,MovieGen,LTX-video,CogVideoX} generate high-quality videos, their resource demands make them unsuitable for on-device use. 

\textbf{On-Device Models.}
In contrast, only limited work targets on-device video generation~\citep{wu2024snapgenvgeneratingfivesecondvideo,kim2025ondevicesoraenablingtrainingfree}. 
The Wan2.1 family~\citep{wan2025wanopenadvancedlargescale} includes a $1.3$B T2V model, but its low VAE compression yields too many latent tokens for deployment. 
LTX-Video~\citep{hacohen2024ltxvideorealtimevideolatent} applies a high-compression VAE and runs in real time on GPUs, yet its $1.9$B parameters remain prohibitive for mobile devices. 
SnapGen-V~\citep{wu2024snapgenvgeneratingfivesecondvideo} adopts a lightweight UNet but sacrifices visual fidelity. 
Mobile Video Diffusion~\citep{yahia2024mobilevideodiffusion} reduces Stable Video Diffusion~\citep{SVD} by pruning channels and blocks. 
On-device Sora~\citep{kim2025ondevicesoraenablingtrainingfree} achieves low-resolution video generation on iPhones via temporal token merging and concurrent block loading to handle memory limits. 

\textbf{Step Distillation.} 
Diffusion models~\citep{stablediffusion3,SDXL,simple-diffusion} require many denoising steps, each involving a full network pass, which creates latency. 
Reducing the number of steps directly improves efficiency. 
Numerous methods address this in text-to-image tasks~\citep{dmd,dmd2,cfm,wang2024rectified,kim2024simple,mei2024codi,dao2025swiftbrush}, representative work including progressive distillation~\citep{salimans2022progressive,li2024snapfusion}, consistency models~\citep{song2023consistency,song2023improved}, adversarial training~\citep{ufogen,add,ladd}, shortcut models~\citep{frans2024stepdiffusionshortcutmodels}, and mean flow~\citep{geng2025mean}. 
For video, \citet{zhang2024sfv,wu2024snapgenvgeneratingfivesecondvideo} achieves few-step generation \citet{SVD} with adversarial training specially-designed spatio-temporal discriminator.

\section{Preliminaries}
\label{sec:preliminaries}
Following popular practices of latent diffusion~\citep{opensora}, we employ a video autoencoder to encode video data $\mathbf{X} \in \mathbb{R}^{3 \times T \times H \times W}$ into a compressed latent space $\mathbf{x} \in \mathbb{R}^{c\times t \times h \times w}$, where $T$ is the number of temporal frames, $H$ and $W$ are the spatial resolutions, and $c$ is the latent channels. 
The VAE compression ratio is thus $\frac{T}{t}\times\frac{H}{h}\times\frac{W}{w}$, e.g., $4\times8\times8$ \citep{CogVideoX,HunyuanVideo,wan2025wanopenadvancedlargescale} and $8\times16\times16$ \citep{CosmosTokenizer} VAEs. 
The objective of the DiT generator is to generate $\mathbf{x}$ under certain guidance (i.e., text prompt). 

We employ Rectified Flow~\citep{wang2024rectified} to train our latent DiT model. 
According to the flow-matching-based diffusion process, given a clean video latent $\mathbf{x}_0=\mathbf{x}$, the intermediate noisy state $\mathbf{x}_t$ at a timestep $t$ is:
\begingroup
\setlength{\abovedisplayskip}{1pt}
\setlength{\belowdisplayskip}{1pt}
\begin{equation}
    \label{equ:forward}
    \mathbf{x}_t = \left(1 - t\right)\mathbf{x}_0 + t \epsilon, \text{where}~\epsilon \sim \mathcal{N}\left(0, \mathit{I}\right),
\end{equation}
\endgroup
which is a linear interpolation between the data distribution and a standard normal distribution. 
The model aims to learn a vector field $v_\theta\left(t, \mathbf{x}_t\right)$ using the Conditional Flow Matching objective, \ie,
\begingroup
\setlength{\abovedisplayskip}{1pt}
\setlength{\belowdisplayskip}{1pt}
\begin{equation}
    \label{equ:loss:fm}
    \mathcal{L}_\text{fm} = \mathbb{E}_{t, \epsilon, \mathbf{x}_0}\left\Vert v_\theta\left(t, \mathbf{x}_t\right) - (\epsilon - \mathbf{x}_0) \right\Vert_2^2.
\end{equation}
\endgroup

%% file: tex/3_method.tex
\input{figure/pruning}

\section{Method}

We optimize Diffusion Transformer~(DiTs) for efficient on-device video generation from four perspectives:
\begin{enumerate*}[label=\bf{(\alph*)}]
    \item \textbf{High compress VAE:} we employ a high-compression autoencoder, as the computational complexity of transformer scales quadratically with token length.
    Reducing the number of token decreases computation while enabling video generation at higher resolution and longer duration.
    \item \textbf{Efficient DiT architecture:} we design an efficient DiT using a KD-Guided Tri-Level pruning method.
    The method balances model fidelity to the baseline with the hardware constraint of the target device.
    \item \textbf{Step-distillation:} we adopt adversarial fine-tuning for step distillation with a new discriminator head design, reducing the number of sampling steps and achieving up to $20\times$ acceleration during inference.
    \item \textbf{Operator Optimization:} we identify the memory bottlenecks in DiT feed-forward layers and introduce a tiled GEMM strategies, alleviating bandwidth limitations and enabling efficient on-device inference without requiring kernel/compiler modification.
\end{enumerate*}

\subsection{Scaling Latent Compression Ratio} \label{sec:quality and efficiency}
DiT demonstrates superior generation capabilities when attending on full token length ($thw$), however, it is also notorious for its quadratic computational cost.
The key idea of the latent diffusion model is to construct a compressed latent space and reduce the generation cost. 
As a result, a straightforward idea to accelerate DiT is to further increase the VAE compression ratio. 
State-of-the-art models (CogVideoX~\citep{CogVideoX},Hunyuan\citep{HunyuanVideo},Wan~\citep{wan2025wanopenadvancedlargescale}) employ a $4\times8\times8$ VAE combined with a $1\times2\times2$ patchify module, which comprises a $4\times16\times16$ total compression rate, while the recent OpenSora-2~\citep{opensora} adopts a $4\times32\times32$ VAE, and LTX~\citep{LTX-video} adopts an $8\times32\times32$ VAE to reduce the dimensionality of the latent features input to the DiT and results in faster generation speed. 
However, there has been limited research on how the VAE compression ratio affects the quality and speed of video generation.
Upon aggressive compression, it becomes more challenging for the VAE decoder to fully reconstruct the details, which may result in quality loss. 
In this work, we perform a comprehensive study on the scaling of the VAE compression ratio. 
We follow \citet{LTX-video,wu2025h3ae} and construct video VAEs with various compression ratios from $4\times16\times16$ to $8\times64\times64$. 
We build the VAE with 3D convolutions to better handle video modality, and use a fixed latent channel number for all variants. 
We train the same DiT network under each latent space and benchmark the generation speed and quality. 
Results and discussions are in~\cref{sec:exp_vae,sec:vae training}. 

\input{figure/sensitivity}

\subsection{Efficient DiT Architecture via KD-Guided Tri-Level Pruning}
\label{sec: pruning & KD}
Despite operating in a highly compressed latent space, the size of the Diffusion Transformer (DiT) remains a critical factor in edge generation scenarios~\citep{wu2024snapgenvgeneratingfivesecondvideo}, where mobile devices are constrained by limited memory, power, and computational resources. Training a compact DiT that still achieves high-quality generation is a non-trivial challenge.
First, DiT models generally exhibit strong generation capabilities only when scaled to a sufficiently large capacity. Moreover, designing an effective small-scale DiT is difficult due to the high-dimensional design space—including network depth (number of transformer blocks), width (channel size), and attention head count.

A promising alternative is to begin with a well-trained large model and prune it to meet resource constraints. Prior work such as TinyFusion~\citep{fang2024tinyfusion} explored this approach via block-wise pruning using a learnable layer mask, effectively constructing a shallower DiT. However, as the properties demonstrated in~\cref{fig: sensitivity}, it is still of great value to design a fine-grained pruning method that offers deeper insight into which parameters are critical or redundant, thereby enabling a better trade-off between efficiency and generation quality.

To address these, we propose a tri-level pruning scheme combined with knowledge alignment, enabling us to derive an efficient DiT architecture from a larger teacher model. Our approach maintains competitive performance while meeting the requirements for edge deployment.


\subsubsection{Tri-level Pruning}
As exhibited in~\cref{fig: sensitivity}, the transformer block pruning is a simple yet coarse approach, we consider it a low-granularity pruning. 
To enable finer granularity and better address redundancies, we propose a tri-level pruning scheme that incorporates block pruning and further introduces fine-grained pruning techniques, including head pruning for the multi-head attention mechanism and channel pruning for the linear layer.
Notably, the pruned model can be converted into a dense and compact form, enabling execution on mobile devices without requiring additional compilation or specialized hardware support.

We employ a set of learnable binary masks to implement the tri-level pruning scheme.
Each binary mask encodes the importance of its corresponding granularity, \ie \emph{block}, \emph{attention-head}, or \emph{linear dimension}.
A mask value of 0 indicates that the corresponding unit should be pruned, while a value of 1 denotes that it should be preserved.
Specifically, block pruning can be formulated as shown in~\cref{equ: block prune}, where $y_{b_i},x_{b_i}$ indicate the input and output features of the $b_i^\text{th}$ DiT block, $m_{b_i} \in \{0, 1\}$ is the binary mask associated with that block, and $\mathcal{M}_b = [m_{b_1}, \dots, m_{b_N}] \in \{0, 1\}^N$ denotes the set of binary masks for all DiT blocks.
When $m_{b_i}=1$, the block is active; otherwise, its output is bypassed through a residual connection.
\begin{equation}
    y_{b_i} = \text{Block}_{b_i}(x_{b_i}) \odot m_{b_i} + x_{b_i} \odot (1 - m_{b_i}), \quad m_{b_i} \in \{0, 1\}, \mathcal{M}_b = [m_{b_1}, \dots, m_{b_N}] \in \{0, 1\}^N
\label{equ: block prune}
\end{equation}
The other two pruning schemes can be expressed using a unified formulation, since pruning \emph{attention heads} is equivalent to removing specific output features before the multi-head attention operation for each token.
By integrating the pruning mechanism into the linear layer, the operation can be formulated as shown in~\cref{equ: linear prune}:
\begin{equation}
    y_{l_i} = \text{Linear}_{i}(x_{l_i}, W_{l_i}, b_{l_i}) \odot \mathcal{M}_{l_i}, \quad m^d_{l_i}\in\{0, 1\}, \mathcal{M}_{l_i}=[m^1_{l_i}, \cdots,m^D_{l_i}]\in \{0, 1\}^D
    \label{equ: linear prune}
\end{equation}
where $y_{l_i},x_{l_i}$ denote the output and input features of the $l_i^\text{th}$ linear layer, and $\mathcal{M}_{l_i}\in \{0, 1\}^D$ is a binary mask with $D$-dimension corresponding to the output channels.
For each $m^d_{l_i} \in {M}_{l_i}$, a value of $m^d_{l_i}=0$ zeros out the corresponding output channel at dimension $d$ for layer $l^\text{th}_i$; otherwise the channel remains active.

The proposed tri-level pruning scheme begins by generating a candidate mask set $\mathbb{M}$ for each pruning target, as illustrated in~\cref{fig: pruning}.
These candidate masks are selected based on the desired number of active components, which are constrained by the memory limitation of the target device.
Notably, exhaustively exploring all pruning combinations results in an extremely large search space, making the optimization problem intractable~(\eg, pruning 6 out of 32 attention heads results in 906,192 possible configurations).
To mitigate this, we adopt a group-wise masking mechanism that partitions overall search space into smaller subspaces, allowing pruning to be performed efficiently within each subspace.
Once the candidate masks are generated, we further optimize them to identify the optimal configuration that minimizes the information loss caused by pruning.

\subsubsection{Knowledge Distillation via Feature Alignment}
\label{sec:KD feature align}

Knowledge distillation~(KD) is a widely adopted technique for transferring knowledge from a teacher model to a student model.
Therefore, it is an effective strategy for preserving the performance of the pruned model.
However, due to varying pruning schemes, the pruned student model may have different feature widths compared to the teacher model, which poses challenges for traditional distillation.
Inspired by~\citet{yu2024repa}, we employ a trainable affine transformation to align the features between the teacher and the student model.
Thus, distillation is then performed using the aligned features.
This process is formally defined in~\cref{equ:loss:feature distill}, where $y_{\text{t}_i}$ and $ y_{\text{s}_i}$ represent the output features of $i^{\text{th}}$ DiT block group for the teacher model and the student model respectively:
\begingroup
\setlength{\abovedisplayskip}{1pt}
\setlength{\belowdisplayskip}{1pt}
\begin{equation}
    \mathcal{L}_{distill}=\frac{1}{N}\sum^N_{i=1} \text{sim} (y_{\text{t}_i}, \mathcal{F}_i(y_{\text{s}_i}; \Theta_i)  )
    \label{equ:loss:feature distill}
;\end{equation}
\endgroup
Here, $N$ denotes the number of DiT block groups, and $\text{sim}(\cdot,\cdot)$ is a similarity alignment function used to match the feature distributions between teacher and student. 
The function $\mathcal{F}_i(\cdot;\Theta_i)$ is an affine transformation parameterized by $\Theta_i$, introduced to align the dimensionality of the student features with that of the teacher.
The overall training loss is formulated as~\cref{equ:loss:all}, where $\mathcal{L}_{\text{flow-matching}}$ is the conditional flow-matching objective from~\cref{equ:loss:fm}, and $\alpha$ is a hyper-parameter to adjust the weight of distillation.
$\alpha$ is set to $0.01$ in our experiments.
\begingroup
\setlength{\abovedisplayskip}{1pt}
\setlength{\belowdisplayskip}{1pt}
\begin{equation}
    \mathcal{L}=\mathcal{L}_{\text{fm}} + \alpha \mathcal{L}_{\text{distill}} 
    \label{equ:loss:all}
\end{equation}
\endgroup

\subsubsection{Integration with Hardware-aware Objective}
Here, we specify the details of our tri-level pruning scheme for constructing an efficient Diffusion Transformer architecture tailored to the iPhone 16 Pro Max.
Due to the memory limitation of the device, the total number of parameters must remain under 1 billion.
Based on the sensitivity analysis in~\cref{fig: sensitivity}, which indicates that FFN contributes more significantly to performance than attention heads, we prioritize pruning attention heads more aggressively.
Starting from a 2B parameter base model with 28 DiT blocks, 32 attention-heads, and FFN dimension of 8192, our final efficient model archives 915M parameters by pruning 8 blocks, 12 attention heads, and reducing the FFN dimension by 25\% following~\cref{algorithm: search prune config}.
More details can be found in~\cref{sec:search algo}.

\subsection{Adversarial Fine-tuning for Step Distillation}
\label{sec:step-distill}
We adopt adversarial fine-tuning of step-distillation, 
following ~\citet{wu2024snapgenvgeneratingfivesecondvideo}, with a generator generator $\mathcal{G}_\theta(t, \mathbf{x}_t)$ and a discriminator $\mathcal{D}\phi(t, \mathbf{x}_t)$. 
The generator $\mathcal{G}_\theta$ is initialized with the pretrained DiT denoiser weights.
The discriminator $\mathcal{D}\phi$ inherits the DiT backbone, whose first $K$ blocks are initialized from $\mathcal{G}_\theta$ and frozen to serve as a timestep-conditioned feature parser, while the subsequent DiT block is learnable and include 3D self-attention and cross-attention to enhance spatio-temporal capacity.
An MLP head with SiLU activation is appended to produce real/fake logits as shown in \cref{fig: pruning}.
The generator $\mathcal{G}_\theta$ learns to generate clean samples in a few steps~(\ie 4-step), while the discriminator $\mathcal{D}\phi$ distinguishes real/generated samples, see \cref{sec:append:step-distill} for details.
The adversarial fine-tuning reduces diffusion sampling budget by up to $20\times$ comparing to full-step diffusion.

\subsection{Tiled GEMM for Efficient FFN Inference}
\label{sec:tiled_gemm}

In Transformers, the Feed-Forward Network~(FFN) is a token-wise two layer MLP applied after the attention mechanism.
It expands the channel dimension from $d$ to $N d$~(typically $N{\in}[2,4]$) and projects back to $d$ via a nonlinearity activation layer~(\eg SiLU).
The design increases the expressive capacity of the token-wise mapping, raising the effective rank and refining token-level representation, while attention primarily mixes information across tokens.

While the expansion in FFN~($d \rightarrow N d \rightarrow d$) improves quality, the large General Matrix to Matrix Multiplications~(GEMMs) become a memory bottleneck on mobile devices.
Although our KD-guided pruning~(\cref{sec: pruning & KD}) and adversarial step distillation~(\cref{sec:step-distill}) reduce overall complexity, GEMMs in FFN remain limited by the device's memory bandwidth.
Addressing this bottleneck with conventional kernel-level optimization is infeasible as the deployment compiler, Apple's CoreML, is a closed-source tool.
Therefore, we introduce an operator-level tiled GEMM strategy for the $d \rightarrow N d \rightarrow d$ projections as illustrated in \cref{fig:gemm tile}.
This method partitions the weight matrix along the expansion dimension $N d$ into smaller, cache-friendly tiles, and activation are fused within each tile to reduce extra reads/writes.
This design mitigates I/O bottleneck, improving cache usage and alleviating bandwidth pressure, particularly for large feature dimensions such as $d{=}8192$.

We benchmark the latency of a fixed number of input tokens ($L{=}2048$) and $N{=}4$ while varying $d$, comparing a na\"ive implementation to the tiled GEMM as shown in \cref{fig:gemm tile bench}.
For reference we also plot the theoretical scaling estimated from FLOPs, $4LNd^2$.
As $d$ increases, the na\"ive GEMM's latency grows faster than the theoretical baseline, indicating a memory-bound issue, whereas the tiled GEMM remains close to the theoretical baseline, demonstrating reduced memory traffic.

The tiled-GEMM in FFN yields an $\sim$10\% end-to-end DiT forward speedup on-device, without fine-tuning and complementary to the KD-guided pruning and step-distillation benefit.

%% file: figure/pruning.tex
\begin{figure}[t]
    \centering
    \includegraphics[width=0.95\linewidth]{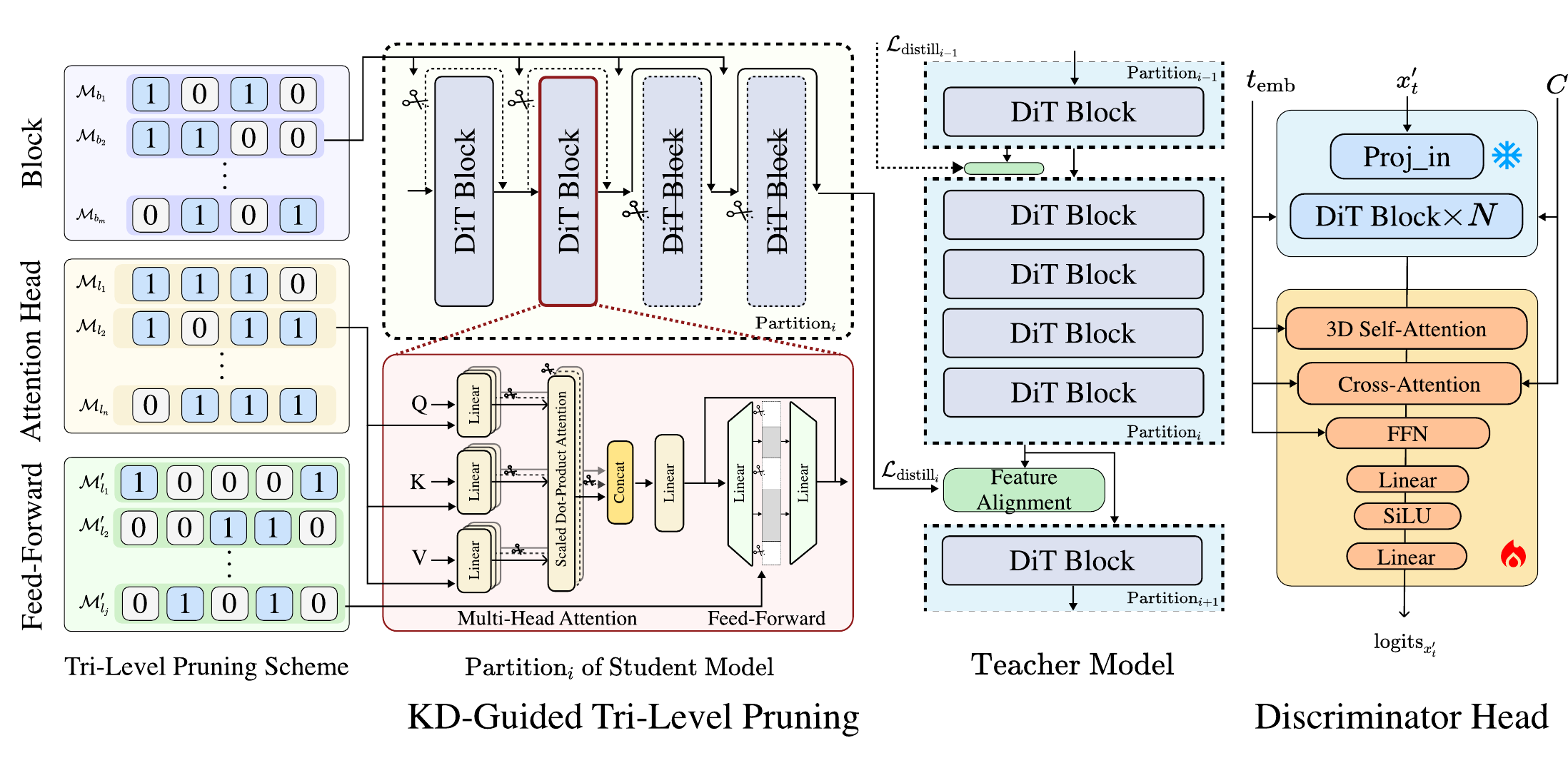}
    \vspace{-1em}
    \caption{\textbf{Overview of proposed KD-Guided Tri-Level Pruning and new discriminator head.}
    The tri-level pruning scheme operates across three levels of granularity, the block, attention-head, and feed-forward network dimension, ranging from coarse to fine. This design enables flexible, efficient, and stable model compression.
    Additionally, the proposed discriminator adopts standard DiT blocks with a MLP classifier head, improved condition alignment for adversarial training.}
    \label{fig: pruning}
    \vspace{-4pt}
\end{figure}

%% file: figure/sensitivity.tex
\begin{figure}[t]
    \includegraphics[width=1.\linewidth]{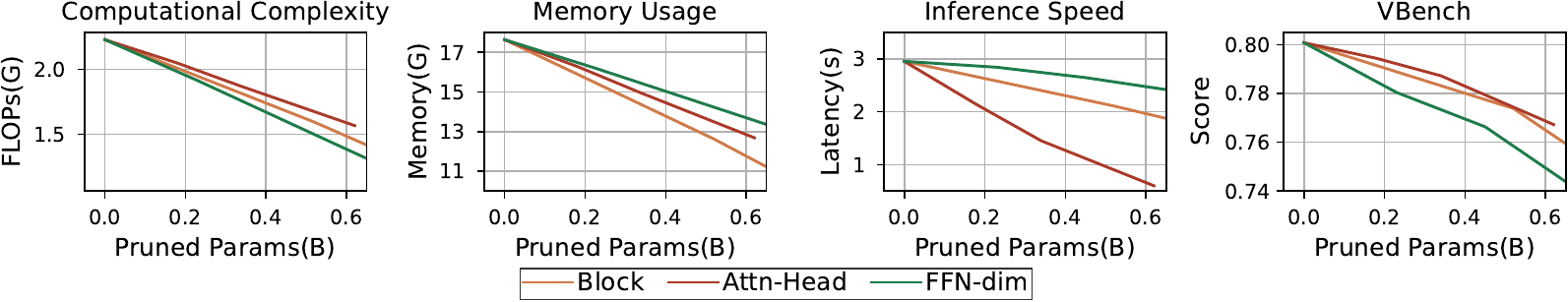}
    \vspace{-1em}
    \caption{Sensitivity Analysis of DiT Components. The sensitivity analysis is conducted by progressively pruning DiT blocks, attention-heads and feed-forward network~(FFN) dimension. For each setting, we benchmark FLOPs, memory usage, inference speed, and VBench score to assess the impact of each component on model efficiency and performance.}
    \label{fig: sensitivity}
    \vspace{-2pt}
\end{figure}

%% file: tex/4_evaluation.tex
\section{Experiments}
\label{sec:experiments}
\noindent\textbf{Training.} We train on both curated real-world image/video data and synthetic data. 
We use $128$ NVIDIA A100 80GB GPUs for DiT training, using AdamW optimizer with $5e-5$ learning rate and betas values as $\left[0.9, 0.999\right]$. 
We build our Diffusion Transformer following public models~\citep{CogVideoX,LTX-video}, and incorporate QK normalizations and Rotary Positional Embeddings (RoPE)~\citep{su2024roformer}. 
The T5 text encoder~\citep{T5} is employed to capture textual information.
The training is conducted using low resolution image and video for pretraining and then finetuning with high resolution data.
More training details in \cref{sec:dit training}.

\noindent\textbf{Adversarial Fine-tuning} is conducted for $20K$ iterations on $64$ A100 GPUs, using the AdamW optimizer with a learning rate of $1e-6$ for the generator (\ie, DiT) and $1e-4$ for the discriminator heads.
We set the betas as $\left(0.9, 0.999\right)$ for both the generator and the discriminator optimizers. We set the EMA rate as $0.95$ following \citet{zhang2024sfv}. 
Additional details are reported in \cref{sec:dit training}.

\noindent\textbf{Evaluation and Deployment.} Our models are evaluated following the standard Vbench~\citep{huang2023vbench} setting, that is, we generate 5 videos for each prompt, and test the scores over the 1K prompt set. 
Both server and mobile-deployed models are step-distilled and evaluated with 4-step generation.
The server model generates $121$-frame horizontal videos at a resolution of $576 \times 1024$, without classifier-free guidance.
The generated video is saved at 5 seconds 24 FPS for score testing and qualitative visualization. 
We use different seeds and find the $\Delta\text{VBench score}$  is lower than $\pm 0.2$. 
For mobile demo, we generate $49\times384\times512$ videos on iPhone 16 Pro Max using CoremlTools \citep{coreml2021} under half precision.
We employ the CLIP text encoder for on-device text encoding efficiency, while the T5 encoder is utilized for the server-side model. 
Additional details are in~\cref{sec:append:mobile-deploy}.
We measure the latency by 50 runs and take the median.


\subsection{Qualitative Results}
\label{sec:exp_visual}
\begin{figure}[t]
  \centering
    \includegraphics[width=\linewidth]{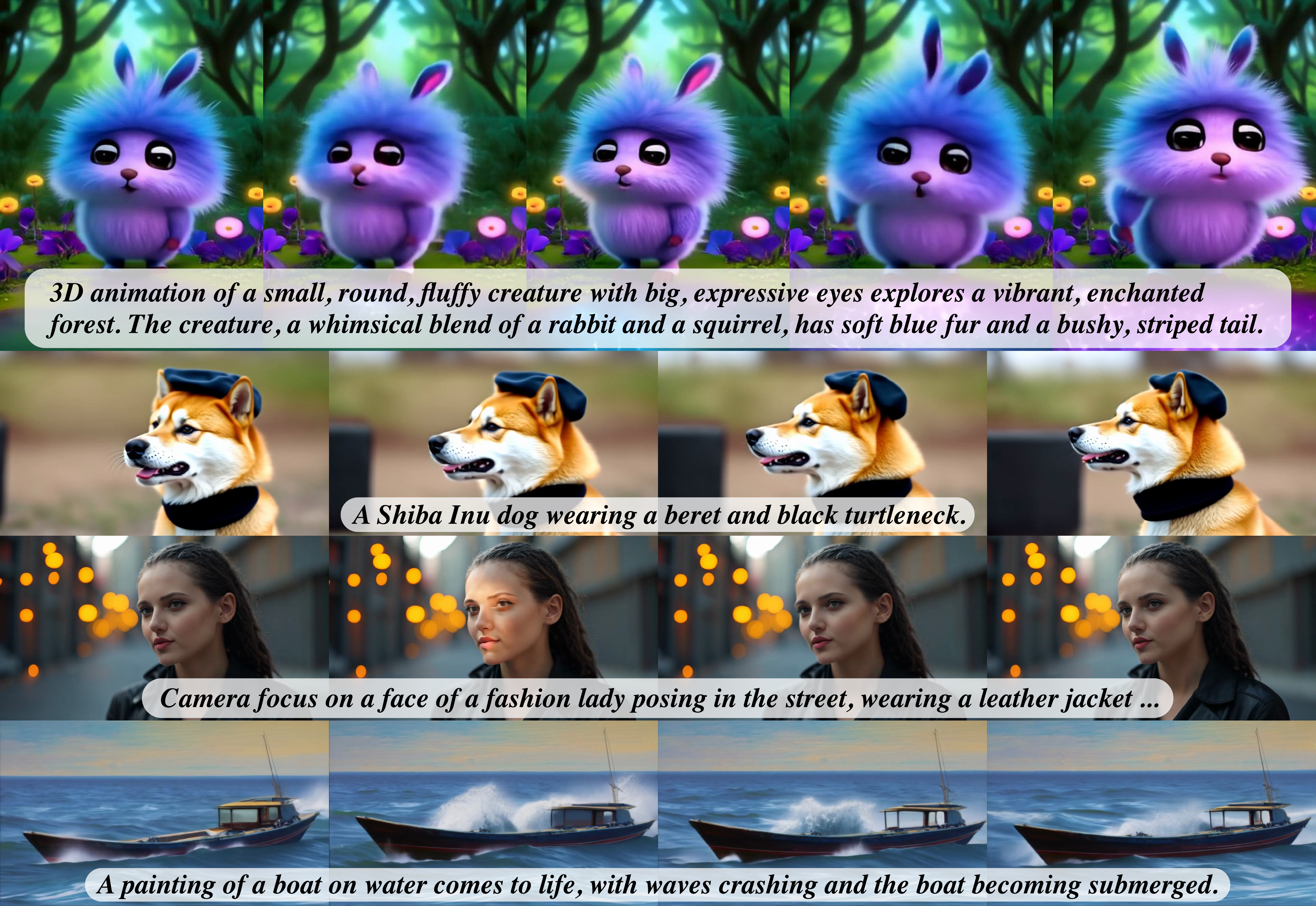}
   \vspace{-1.5em}
   \caption{Video generated by our efficient diffusion transformer.}
   \label{fig:video}
\end{figure}

We visualize our generated videos in \cref{fig:video}. Our model consistently produces high-quality video frames and smooth object movements. 
To demonstrate the generic text-to-video generation ability, we show various generation examples, including human, animal, photorealistic, and art-styled scenes. 
We include more video visualizations and the efficient model demo in \emph{supplementary material}.


\input{table/vbench}

\subsection{Quantitative Benchmark}
\label{sec:exp_quant}

We evaluate our method on VBench~\citep{huang2023vbench} and compare it against recent state-of-the-art DiT-based video generation models, as shown in \cref{table: vbench comparison}.
Although our model is compact and optimized for fast inference on mobile devices, 
it achieves a higher total score than several strong baselines, including the OpenSora-1.2, CogVideoX-2B~\citep{CogVideoX}, LTX-Video~\citep{LTX-video}.
Compared to current open-source SOTA, Wan2.1-1.3B~\citep{wan2025wanopenadvancedlargescale}, our server variant achieves comparable quality while delivering faster inference speed per sampling step. 
Importantly, the mobile deployment ~(0.9B parameters) maintains competitive scores relative to larger models while running efficiently on the iPhone 16 Pro Max. 
These results highlight the effectiveness of our DiT pruning and distillation method.
Human evaluation studies further demonstrate the perceptual quality of our models as reported in \cref{sec:user study}.
Comparison between our mobile variant and other mobile efficient methods are shown in \cref{sec:comp on-device-model}.


\subsection{Ablation Study}

\noindent\textbf{Scaling VAE Compression Ratio.}
\label{sec:exp_vae}
In \cref{table:vae}, we scale the VAE compression ratio and compare DiT generation speed and video quality. 
For a fair comparison, we train a 2B-parameter DiT with each VAE and measure per-step generation speed by testing one denoising step on the Nvidia A100 GPU at $121\times576\times1024$ resolution. 
We observe that though lower compression VAEs ($4\times16\times16$) can achieve better reconstruction PSNR, the generation speed is slower by magnitudes. 
On the other hand, aggressive compression ($8\times64\times64$) results in poor reconstruction, and will negatively impact generation quality~(\ie VBench scores). 
We find that ($8\times32\times32$) hits a balance between speed and quality, and employ this configuration for our Diffusion Transformer. 
The training details and more experiment results for the video VAE can be found in \cref{sec:vae training,sec:more vaes}.

\input{table/vae}

\noindent\textbf{Impact of KD-Guided Training.}
We evaluate the effect of the proposed KD-guided training via ablation with and without distillation.
As shown in~\cref{tab:ablate:prune method}, the tri-level trained with the proposed distillation objective consistently outperforms the counterpart without distillation, indicating that it helps recover capacity lost due to pruning.

\noindent\textbf{Impact of Tri-Level Pruning.}
We compare the proposed tri-level pruning against random masking and block-only pruning (\emph{shallow}) following \citet{fang2024tinyfusion,xiesana1.5}. 
All variants are initialized from the same pretrained 2B DiT.
As shown in~\cref{tab:ablate:prune method}, the tri-level pruned model consistently outperforms both random-masked and \emph{shallow} baseline in Quality and Total scores.
These results indicate the proposed tri-level pruning can effectively remove redundant components while preserving capacity, thereby achieving pruning with minimum performance degradation.


\noindent\textbf{Comparison with Compact Model Trained from Scratch.}
To separate the effect of tri-level pruning and KD-guided training from model size, we train a compact DiT the same model configuration from scratch.
As reported in \cref{tab:ablate:prune method}, this compact baseline underperforms the KD-guided tri-level pruned model by large margin across Quality, Semantic, and Total scores.
These results indicate that training a small DiT from scratch is suboptimal, while starting from a larger teacher and applying tri-level pruning with knowledge distillation is crucial for preserving overall quality under mobile constraints.

\input{minipages/ablation_tables_1}

\noindent\textbf{Full Guidance Adversarial Distillation. }
We validate the proposed discriminator tailored for the DiT denoiser through an ablation study on the prediction head. 
The experiments are conducted using our pre-trained 2B parameter DiT model with various discriminator head designs.  
We compare our design with spatial-temporal heads introduced in~\citet{wu2024snapgenvgeneratingfivesecondvideo} and the lightweight ResBlock head proposed in~\citet{wang2024phased}.
Since the discriminator head in~\citet{wang2024phased} was originally designed for the text-to-image model, we extend it to a Conv3D variant in the ablation.
We show that the proposed prediction head~(a transformer block followed by an MLP classifier) achieves best 4-step generation performance, with notable gains in semantic scores.
We attribute this to improved alignment between the text condition and the hidden states.

%% file: table/vbench.tex
\begingroup
\setlength{\textfloatsep}{0pt}
\setlength{\intextsep}{0pt}
\begin{table*}[t]
\small
\centering
\caption{VBench~\citep{huang2023vbench} comparison with popular open-source Diffusion Transformer video generation models. Scores for open-source models are collected from the~\href{https://huggingface.co/spaces/Vchitect/VBench_Leaderboard}{VBench Leaderboard}.}
\vspace{-1em}
\resizebox{\linewidth}{!}{
\renewcommand\arraystretch{1}
\renewcommand\tabcolsep{1.5pt}
\begin{tabular}{l|c|ccc|cccccc}
\toprule
Model & Params (B) & Total & Quality & Semantic & Flickering & Aesthetics & Imaging & Obj. Class & Scene & Consistency \\
\midrule
Wan2.1        & 14 & 84.70 & 85.64 & 80.95 & 99.53 & 61.53 & 67.28 & 94.24 & 53.67 & 27.44 \\
Wan2.1        & 1.3  & 83.31 & 85.23 & 75.65 & 99.55 & 65.46 & 67.01 & 88.81 & 41.96 & 25.50 \\
Open-Sora-2.0 & 11 & 84.34 & 85.40 & 80.12 & 99.40 & 64.39 & 65.66 & 94.50 & 52.71 & 27.50 \\
Open-Sora-1.2 & 1.2 & 79.76 & 81.35 & 73.39 & 99.53 & 56.85 & 63.34 & 82.22 & 42.44 & 26.85 \\
Hunyuan       & 13 & 83.24 & 85.09 & 75.82 & 99.44 & 60.36 & 67.56 & 86.10 & 53.88 & 26.44 \\
CogVideoX1.5  & 5  & 82.01 & 82.72 & 79.17 & 98.53 & 62.07 & 65.34 & 83.42 & 53.28 & 27.42 \\
CogVideoX     & 5  & 81.91 & 83.05 & 77.33 & 78.97 & 61.88 & 63.33 & 85.07 & 51.96 & 27.65 \\
CogVideoX     & 2  & 81.55 & 82.48 & 77.81 & 98.85 & 61.07 & 62.37 & 86.48 & 50.04 & 27.33 \\
Step-Video    & 30 & 81.83 & 84.46 & 71.28 & 99.40 & 61.23 & 70.63 & 80.56 & 24.38 & 27.12 \\
Mochi-1       & 10 & 80.13 & 82.64 & 70.08 & 99.40 & 56.94 & 60.64 & 86.51 & 36.99 & 25.15 \\
LTX-Video     & 1.8  & 80.00 & 82.30 & 70.79 & 99.34 & 59.81 & 60.28 & 83.45 & 51.07 & 25.19 \\
\hline
Ours-Server   & 2.0  & 83.09 & 84.65 & 76.86 & 98.74 & 64.72 & 65.85 & 90.57 & 52.76 & 27.28 \\
Ours-Mobile   & 0.9  & 81.45 & 83.12 & 74.76 & 98.11 & 64.16 & 63.41 & 92.26 & 51.06 & 25.51 \\ 
\bottomrule
\end{tabular}
}
\label{table: vbench comparison}
\end{table*}
\endgroup

%% file: table/vae.tex

\begingroup
\setlength{\textfloatsep}{0pt}
\setlength{\intextsep}{0pt}
\begin{table*}[t]
\small
\centering
\caption{Scaling VAE compression ratio. VAE PSNR is measured on DAVIS~\citep{DAVIS} with $33\times512\times512$ resolution. Latencies are for one denoising step.
VBench scores are provided.
}
\vspace{-1em}
\renewcommand\arraystretch{1}
\renewcommand\tabcolsep{3pt}
\resizebox{0.85\linewidth}{!}{%
\begin{tabular}{@{}cc c cccccc@{}}
\toprule
\multicolumn{2}{c}{\textbf{VAE}} & \multicolumn{7}{c}{\textbf{Diffusion Transformer}} \\
\cmidrule(r){1-2} \cmidrule(l){3-9}
\textbf{Compression Factor} & \textbf{PSNR} & \textbf{Latency~(ms)} &
\textbf{Total} & \textbf{Quality} & \textbf{Semantic} & \textbf{Aesthetic} & \textbf{Consistency} & \textbf{Flickering} \\
\midrule
$4\times16\times16$ & 33.1 & 7900 & 80.35 & 82.05 & 73.54 & 64.45 & 26.80 & 98.59 \\
$4\times32\times32$ & 30.9 &  920    & 79.95 & 82.99 & 67.83 & 61.52 & 27.07 & 97.46 \\
$8\times32\times32$ & 30.6 &  380    & 79.80 & 82.59 & 68.66 & 61.80 & 27.17 & 97.70 \\
$8\times64\times64$ & 28.2 &   90    & 78.40 & 81.79 & 64.86 & 55.29 & 26.11 & 97.52 \\
\bottomrule
\end{tabular}
}
\label{table:vae}
\end{table*}
\endgroup

%% file: minipages/ablation_tables_1.tex
\vspace{-0.5em}
\begin{minipage}[t]{0.475\textwidth}
\centering
\captionof{table}{Ablation study on tri-level pruning schemes and fine-tuning using proposed knowledge distillation.}
\resizebox{0.9\linewidth}{!}{
\begin{tabular}{c c c | ccc}
\toprule
\textbf{Method} & \textbf{KD} & \textbf{Params(M)} & \textbf{Quality} & \textbf{Semantic} & \textbf{Total}\\
\midrule
tri-level & \ding{51} & 915  & 83.12 & 74.76 & 81.45 \\
tri-level & \ding{55} & 915  & 82.19 & 66.23 & 79.00 \\
random & \ding{55} & 915  & 82.01 & 65.01 & 78.68 \\
\emph{shallow} & \ding{55} & 932 & 81.63 & 67.15 & 78.73 \\
compact\textsuperscript{1} & \ding{55} & 915 & 79.23 & 63.94 & 76.17 \\ 
\bottomrule
\multicolumn{6}{l}{\textsuperscript{1} Train from scratch}
\end{tabular}
}
\label{tab:ablate:prune method}
\end{minipage}
\hfill
\begin{minipage}[t]{0.475\textwidth}
\centering
\captionof{table}{Ablation study on different discriminator head design. The evaluation is conducted with a 4-step generation without classifier-free guidance.}
\resizebox{\linewidth}{!}{
\renewcommand\arraystretch{1}
\renewcommand\tabcolsep{2pt}
\begin{tabular}{lc|ccc}
\toprule
\textbf{Head} & \textbf{\#Steps} & \textbf{Quality} & \textbf{Semantic} & \textbf{Total} \\
\midrule
DiT block + MLP & 4 & 83.81 & 72.89 & 81.63 \\
ResBlock-2D + Temporal-Attn & 4 & 83.24 & 67.78 & 80.14 \\
Lightweight ResBlock & 4 & 80.05 & 66.01 & 77.24  \\

\bottomrule
\end{tabular}
}
\label{tab:ablate:discriminator head}
\end{minipage}

%% file: tex/5_conclusion.tex
\section{Conclusion}
\vspace{-0.5em}
\label{sec:conclusion}
In this work, we present an efficient video generation framework that significantly accelerates Diffusion Transformers, making efficient synthesis feasible on mobile devices. 
By combining a high-compression VAE, latency- and sensitivity-aware pruning, and adversarial step distillation, we successfully deploy DiT video generator to iPhone and reduce inference to just four steps while maintaining high visual quality. 
Our pipeline achieves over $15$~FPS generation speed (generate 49-frame within 4 seconds) on an iPhone 16 Pro Max, demonstrating the practical viability of DiT-based video generation on edge devices. 
We discuss limitations and broader impact in \cref{sec:append:limitation and boader impact}.

%% file: tex/appendix.tex

\section{User Study}
\label{sec:user study}
To evaluate the human preference across different video generation models, we conduct human evaluations comparing our model against baselines such as CogVideoX-2B, LTX-Video in \cref{tab:user study}.
We generate video clips using prompts from VBench and MovieGenBench~\citep{MovieGen} and
ask human labelers to select the best results across prompt alignment, aesthetics, and motion quality.
The results indicate that our model significantly outperforms the baselines.

To further demonstrate the effectiveness of our proposed method, we also evaluate the human preferences across our server-side model and mobile-deployed model in \cref{tab:user study ours}.
We generate video samples in $49\times512\times384$ for both models.
The result illustrates the trade-off in visual quality in terms of efficiency.
Notably, our mobile-deployed model uses CLIP as text-encoder for efficiency while the server-side model uses T5-Encoder.
\input{table/user-study}

\section{Comparison with Other Mobile Video Generation Methods}
\label{sec:comp on-device-model}
We compare the VBench score of our model against SnapGen-V~\citep{wu2024snapgenvgeneratingfivesecondvideo} and on-device Sora~\citep{kim2025ondevicesoraenablingtrainingfree}, using benchmark metrics reported in their paper to show the performance of our mobile-deployed model as in \cref{table:vbench-mobile}.
Notably, SnapGen-V is based on UNet architecture for efficient video generation and on-device Sora is a training-free method that enables open-sora~\citep{opensora} on the mobile device.
Due to on-device Sora’s published precision and scale differ from the standard VBench values, we have converted them to a common scale and also include its baseline, OpenSora v1.2 as reference.
Note that our mobile model outperformed SnapGen-V by a notable margin in both Semantic Score and Total Score, and it also achieves better performance than the baseline of on-device Sora.
These results demonstrate that our approach provides competitive or superior performance compared to existing mobile video generation methods.

\input{table/vbench_mobile}

\section{Search for Optimal Pruning Configuration}
\label{sec:search algo}
We describe the detail to determine the optimal pruning configuration for on-device model as \cref{algorithm: search prune config}.
Starting from a pretrained DiT, our objective is to satisfy the device memory budget while minimizing quality degradation.

\noindent\textbf{Search space and constraints.}
We enumerate candidate configurations over three granularities:
$\text{Attn-Heads} \in \{20, 24, 28, 32\},
\text{Blocks} \in \{16, 20, 24, 28\},
\text{FFN dim} \in \{5120, 6144, 7168, 8192\}$.
Here, values denote the \emph{kept} attention heads per block, transformer blocks, and FFN dimension, respectively. 
To reduce the search space, we ignore extreme cases (\eg pruning $>50\%$ of blocks).

\noindent\textbf{Sensitivity-guided pruning order.}
Given by the sensitivity analysis in \cref{fig: sensitivity}, we rank pruning sensitivity as:
$\text{FFN dim}>\text{Blocks}>\text{Attn-Heads}$.
Accordingly, we prefer to prune more aggressively on attention heads and conservatively on FFN dimension, with blocks in between.

For each candidate $c$, we initialize all possible binary masks and optimize them with the KD-guided tri-level pruning objectives \cref{equ: block prune,equ: linear prune,equ:loss:all}. 
We employ Gumbel-Softmax sampling for differentiable mask selection.
Candidates are evaluated on a fixed validation set using the average flow-matching objective~\cref{equ:loss:fm} over multiple timesteps~(\ie 25-step), subject to device budget constraints.

\input{algo/search}
\section{Adversarial Finetuning for Step-Distillation}
\label{sec:append:step-distill}
To obtain a $k-$step distillation procedure, we predefine the intermediate diffusion timesteps as $\mathcal{T} = \{T_1, T_2, \dots, T_k\}$ with the following ordering $T_1=1 > T_2 > \dots > T_k > 0$. Typically, $k$ is set to $4$ to achieve a $4-$step diffusion model. Given a real data sample $\mathbf{X}_0$, we can obtain the latent $\mathbf{x}_0$ using the VAE. We sample two timesteps $t$  and $t^\prime$ uniformly at random from the set $\mathcal{T}$ such that $t^\prime < t$. We can construct the fake and real samples using the diffusion forward \cref{equ:forward} and velocity from the generator $\mathcal{G}_\theta(t, \mathbf{x}_t)$ as follows: 
\begin{equation}
    \text{Fake : } \hat{\mathbf{x}}_{t^\prime} = \mathbf{x}_t + \left(t^\prime - t\right) \cdot \mathcal{G}_\theta\left(t, \mathbf{x}_t\right); \,\,\,\, \text{Real : } \mathbf{x}_{t^\prime} = \left(1 - t^\prime\right)\mathbf{x}_0 + t^\prime \epsilon; \epsilon\sim \mathcal{N}(0,I)
\end{equation}

Using the above real and fake samples, we can define the discriminator and generator losses. Below, we employ the widely used~\citep{add, sauer2023stylegan, sauer2021projected, zhang2024sfv} hinge loss~\citep{lim2017geometric} as the adversarial training objective. 
The discriminator's goal is to differentiate between real and fake samples by minimizing:
\begin{equation}
    \label{equ: discriminator adv}
    \begin{split}
    \mathcal{L}_\text{adv}^\mathcal{D} = & 
    \mathbb{E}_{t^\prime, \mathbf{x}_0} \left[\text{ReLU}(1 + \mathcal{D}_\phi\left(\mathbf{x}_{t^\prime}, t^\prime\right)) \right] +  \mathbb{E}_{t, t^\prime, \mathbf{x}_0}\left[\text{ReLU}\left(1 - \mathcal{D}_\phi\left(\hat{\mathbf{x}}_{t^\prime}, t^\prime\right)\right)\right],
    \end{split}
\end{equation}

The adversarial objective for the generator $\mathcal{L}_\text{adv}^\mathcal{G}$ and the reconstruction objective $\mathcal{L}_{\text{recon}}$ are defined as:
\begin{equation}
    \label{equ: generator adv}
    \mathcal{L}_\text{adv}^\mathcal{G} = 
    \mathbb{E}_{t, t^\prime, x_0}[ \mathcal{D}_\phi\left(\hat{\mathbf{x}}_{t^\prime}, t^\prime\right)] ; \,\,\,\,\,\,\,  \mathcal{L}_{\text{recon}} = \sqrt{\left\Vert \hat{\mathbf{x}}_{0} - \mathbf{x}_0 \right\Vert_2^2 + c^2} - c.
\end{equation}
where $\hat{\mathbf{x}}_{0} = \mathbf{x}_t  - t \cdot \mathcal{G}_\theta\left(t, \mathbf{x}_t\right)$, and $c>0$ is an adjustable constant. Following~\citet{zhang2024sfv,hu2024snapgen}, we also incorporate a reconstruction objective to enhance training stability.

\section{Training Details for DiT}
\label{sec:dit training}
\paragraph{Pre-training.}
The DiT training is trained on the internally collected dataset containing high-quality images and videos, which are similar to public large-scale datasets such as~\citet{Panda70M}.
Training is performed in two stages: (i) pretraining on low-resolution (288p) images and videos for $150K$ iterations, followed by (ii) fine-tuning for an additional $50K$ iterations on a mixed-resolution setting (288p and 576p).
The KD-guided tri-level pruning is only conducted to obtain the mobile variant after the pretraining stage.
The fine-tuning stage for the mobile variant takes $50K$ iterations.
Video clips the mobile variant uses is 49-frame clips.
All video clips are resampled to 24 fps and cropped to 5-second segments.
In the first stage, we only adopt T5 encoder as the text-encoder, leveraging its stronger capacity for modeling long caption and capturing richer text information.
During fine-tuning, we additionally incorporate the CLIP text-encoder alongside with T5, since CLIP is the text encoder deployed on-device.
Each encoder output is first projected into the DiT latent space, then the projected embeddings are concatenated and fed into the DiT as conditioning.
To improve robustness, we randomly mask either the T5 embeddings or the CLIP embeddings during training, enabling better model generalization capacity under both text-encoders.

\paragraph{KD-guided Tri-level Pruning.}
The tri-level pruning procedure is initialized from the fine-tuned DiT model.
To determine the pruning ratios at each granularity~(blocks, attention heads, and FFN dimensions), we consider both the hardware constrain of the iPhone 16 Pro Max ~(parameter budget $<1\text{B}$) and the sensitivity analysis in~\cref{fig: sensitivity} and conduct \cref{algorithm: search prune config}.
The optimization is run for $20K$ iterations per candidate configuration to obtain a stable pruning scheme that minimizes performance degradation.
At the end, We also perform knowledge distillation alone without pruning, where the student is distilled directly from the baseline model, to enhance the performance.

\paragraph{Adversarial Fine-tuning.}
The adversarial fine-tuning is applied to both our server and mobile variants.
The generator and discriminator are initialized from the pre-trained model weights.
We use a frozen server variant model to generate with classifier-free guidance scale 5 to produce reconstruction targets.
During the fine-tuning, the hinge loss is utilized as the adversarial objective, while the reconstruction loss is defined by the $l_2$-Norm between the v-prediction of the generator and the frozen model.

\input{minipages/tiled_gemm}
\section{Implementation Details in tiled GEMM}
\label{sec:append:gemm impl}
As introduced in \cref{sec:tiled_gemm}, Linear Layer with large input feature dimensions tend to become memory-bound.
In our implementation, we apply tiled GEMM specifically to the $Nd\rightarrow d$ projection layer in the FFN, which we find to be the dominant bandwidth bottleneck. 
Other linear layers, such as QKV projections in attention, have much smaller hidden dimensions and show only marginal improvements with tiling, thus we retain their standard implementation.

In pratical, we set the number of partition to $k=4$, which we identify as a practical balance between parallelism and cache locality on the iPhone 16 Pro Max.
This operator-level strategy allows us to reduce data traffic without requiring changes to CoreML's kernel backend, making it directly hardware friendly on mobile device.

\section{Training Details for VAE}
\label{sec:vae training}
The VAE is trained on the internal dataset using 64 NVIDIA A100 80GB GPUs.
The model is optimized using the AdamW optimizer with $1e-4$ learning rate, $\beta=(0.9, 0.999)$, and trained with a batch size of 16 per GPU.
The training process including two-stage: we first pretrain the model on $256\times256$ cropped $17$-frame clips for $100K$ iterations, then fine-tuning on a range of resolutions and clip lengths for additional $50K$ iterations.

The VAE is optimized using a combination of a reconstruction loss between the input data and the reconstruction output, and KL-divergence regularization that enforces the latent distribution to follow a normal distribution, as defined in~\cref{equ:vae_loss}.
The KL weight is set to $\lambda=1e-6$.

\begin{equation}
\label{equ:vae_loss}
\begin{aligned}
\mathcal{L}_{\text{recon}} & = \lVert \hat{y} - y \rVert \\
\mathcal{L}_{\text{KL}}    & = \tfrac{1}{2}\!\left[-\log(\sigma^2) - 1 + \sigma^2 + \mu^2\right] \\
\mathcal{L}                & = \mathcal{L}_{\text{recon}} + \lambda \mathcal{L}_{\text{KL}}
\end{aligned}
\end{equation}

\section{More Experiments on VAE Variants}
\label{sec:more vaes}
We further analyze the impact of the number of latent channel to the VAE and the corresponding DiT performance, as shown in \cref{tab:vae latent}.
As expected, increasing latent channels is always beneficial for VAE reconstruction quality.
However, the benefit quickly saturates, \ie for higher compression VAEs, the reconstruction PSNR can not match that of low-compression VAEs simply by increasing latent channels.
Moreover, larger latent channels can harm or destabilize diffusion quality, as display in the $\left|z\right|=512$ setting.
Therefore, we select latent channel for each compression setting according to the diffusion quality~(\ie VBench score).

Beyond the number of latent channels, we also investigate the impact of patchify options on the overall data dimension compression strategy.
Here, we compare direct compression of video data against those use lower-compression VAEs combined with patchify operations later, as shown in \cref{tab:vae variant}.
The latter approach achieves higher reconstruction PSNR due to compression is applied more conservatively along the spatial and temporal dimensions.
For a fair comparison, we keep the number of latent channels same for overall compression strategy after patchify. 

The reconstruction PSNRs are evaluated on DAVIS with $33\times512\times512$ resolution, while VBench score are evaluated following the standard setup.

\input{table/vae_variants}

\section{Impact of Inference Steps}

Our distilled model supports generation with reduced number of inference steps.
We further exhibits the impact of the inference steps for the generation results in~\cref{tab:distill-step}.
\input{table/distill_steps}

\section{Finetune LTX-Video on Internal Datasets}
\label{sec:supp:ft-ltx}
To mitigate the potential impact of training with internal datasets~\cref{sec:conclusion}, we trained the LTX-Video model using our internal dataset, and the VBench score is shown in~\cref{tab:ltx-ours}.
The results demonstrate that our internal dataset achieves on-par performance comparing to the original LTX-Video dataset.

\begin{table}[ht!]
    \centering
    \caption{The VBench score of official LTX-Video and trained using our internal datasets.}
    \resizebox{0.7\linewidth}{!}{
    \begin{tabular}{l|ccc|ccc}
        \toprule
        Model & Total & Quality & Semantic & Aesthetics & Scene & Consistency \\
        \midrule
        LTX-Video         & 80.00 & 82.30 & 70.79 & 59.81 & 83.45 & 25.19 \\
        LTX-Video~(Ours)  & 80.35 & 82.05 & 73.54 & 64.45 & 37.08 & 26.80 \\
        \bottomrule
    \end{tabular}
    }
    \label{tab:ltx-ours}
\end{table}

\section{Mobile Deployment}
\label{sec:append:mobile-deploy}
We deploy our model on an iPhone 16 Pro Max by converting to FP16 and executing on the Neural Engine and the CPU cores.
To improve on-device numerical stability, we adopt HardSiLU as the activation function and LayerNorm for normalization.
For text encoding, we employ the CLIP text encoder for on-device efficiency, while the T5 encoder is still utilized for the server-side model.

\section{Latency Benchmark Results on Mobile}
We provide screenshots illustrating the latency of our mobile model.
The latency is benchmarked on an iPhone 16 Pro Max using Apple's CoreML toolkits within Xcode.
The reported latency corresponds to the median value collected across multiple runs.
The model is chunked to two chunks for efficient loading and inference.
As shown in~\cref{fig:coreml}, the inference time for one-step DiT model is $668.02$~ms in total.
Accordingly, our 4-step model requires $3,021$~ms to generate a 49-frame video at $512\times384$ resolution.

The latency breakdown for each components in the generation pipeline is shown in the \cref{tab:latency breakdown}.
Latency for module loading and diffusion backward process are included in I/O and Misc.
The overall latency for the whole pipeline is $3,318$ ms, resulting in an average generation speed of $15$~FPS.

\input{table/latency_breakdown}
\input{figure/xcode_benchmark}


\section{Use of LLMs}
We used large language models~(\eg ChatGPT, Gemini) solely to assist with manuscript formatting. 
No part of the research design, experimental implementation, or analysis relied on LLMs.

\section{Limitations and Broader Impact}
\label{sec:append:limitation and boader impact}
Despite these advances, our method has several limitations. 
First, the highly compressed latent space and DiT pruning lead to occasional degradations in fine-grained details, particularly in fast motion or complex texture scenes. 
Second, due to various practical constraints, most state-of-the-art video diffusion models (VDMs) used for comparison in this work, including our own, are trained on internally collected video datasets that cannot be fully disclosed or released. 
As a result, direct comparisons may not be entirely fair and reproducible. To mitigate this limitation, we include a reproduction of the LTX model trained on our dataset and report the results in the \cref{sec:supp:ft-ltx}.
This work enables efficient video generation on mobile devices, but also carries the potential risk of misuse for generating fake or inappropriate content.

\section{More Qualitative Results}
We illustrate more qualitative results of video clips generated by our model in~\cref{fig:more server results,fig:more mobile results}.
\input{figure/more_results}

%% file: table/user-study.tex
\begin{table}[ht]
\begin{minipage}[t]{0.48\textwidth}
    \centering
    \captionof{table}{Human preference across different video generation models.}
    \resizebox{\linewidth}{!}{
    \begin{tabular}{l|ccc}
        \toprule
        Model & Prompt Alignment & Aesthetics & Motion Quality \\
        \midrule
        LTX-Video\textsuperscript{*} & 16.7\% & 10.0\% & 6.7\% \\
        CogVideoX-2B & 40.0\% & 33.3\% & 43.3\% \\
        Ours & 43.3\% & 56.7\% & 50.0\% \\
        \bottomrule
        \multicolumn{4}{c}{\footnotesize\textsuperscript{*}We notice the performance of LTX-Video highly depends on the prompts enhancement.} \\
    \end{tabular}
    }
    \label{tab:user study}
\end{minipage}
\hfill
\begin{minipage}[t]{0.48\textwidth}
    \centering
    \captionof{table}{Human preference across our sever-side and mobile-deployed model.}
    \resizebox{\linewidth}{!}{
    \begin{tabular}{l|ccc}
        \toprule
        Model & Prompt Alignment & Aesthetics & Motion Quality \\
        \midrule
        Server-side & 58.8\% & 52.9\% & 55.8\% \\
        Mobile-deployed & 41.2\% & 47.1\% & 44.1\% \\
        \bottomrule
    \end{tabular}
    }
    \label{tab:user study ours}
\end{minipage}
\end{table}


%% file: table/vbench_mobile.tex
\begin{table}[!ht]
\small
\centering
\caption{VBench~\citep{huang2023vbench} comparison with popular open-source Diffusion Transformer video generation models. Scores for open-source models are collected from the~\href{https://huggingface.co/spaces/Vchitect/VBench_Leaderboard}{VBench Leaderboard}.}
\resizebox{\linewidth}{!}{
\renewcommand\arraystretch{1}
\renewcommand\tabcolsep{5pt}
\begin{tabular}{l|ccc|ccccccc}
\toprule
\multirow{2}{*}{Model} & Total & Quality & Semantic & Subject & Background & Temporal & Motion & Dynamic & Aesthetic & Imaging \\
 & Score & Score & Score & Consistency & Consistency & Flickering & Smoothness & Degree & Quality & Quality \\
\midrule
Ours-Mobile   & 81.45 & 83.12 & 74.76 & 95.73 & 96.64 & 98.11 & 99.23 & 58.33 & 64.16 & 63.41\\ 
SnapGen-V & 81.14 & 83.47 & 71.84 & -- & -- & 99.37 & -- & 51.11 & 62.19 & -- \\
Open-Sora V1.2\textsuperscript{*} & 79.76 & 81.35 & 73.39 & 96.75 & 97.61 & 99.53 & 98.50 & 42.39 & 56.85 & 63.34 \\
On-device Sora\textsuperscript{\dag} & -- & -- & -- & 96.00 & 97.00 & 99.00 & 99.00 & 27.00 & 47.00 & 53.00 \\
\bottomrule
\multicolumn{11}{l}{\footnotesize\textsuperscript{*}We report the result of open-sora for reference since on-device Sora reported result precision and scale differ from the standard VBench values.} \\
\multicolumn{11}{l}{\footnotesize\textsuperscript{\dag}We converted reported on-device Sora's VBench values to common scale for comparison.} \\
\end{tabular}
}
\label{table:vbench-mobile}
\end{table}

%% file: algo/search.tex
\begin{algorithm}[t]
\small
\caption{Search for Optimal Pruning Configuration}
\label{algorithm: search prune config}
\begin{algorithmic}[1]

\Require
{
\\
$\hat{\epsilon}_{\theta}^{\text{super}}$: Pretrained DiT. \\
$\mathbb{T}(\cdot)$: A lookup table mapping a configuration $c$ to its parameter count $\#\text{Params}$.\\
$\mathcal{A}$: The set of possible pruning actions (\ie prune 4 block, 4 attention heads, and 12.5\% FFN dimensions).\\
$D_{\text{val}}$: A fixed validation dataset for evaluating $\mathcal{L}_{\text{MSE}}$.\\
$P_{\text{target}}$: The target parameter count for the final subnet (\ie 1B parameters).
}

\Ensure The optimal subnet configuration: $C_{\text{optimal}}$

\Statex
\State $\rightarrow$ \textbf{Joint search for an optimal subnet:}
\State Initialize current configuration $C_{\text{current}} \gets C_{\text{super}}$ (the largest architecture)

\While{$\mathbb{T}(C_{\text{current}}) > P_{\text{target}}$}
    \State \textit{// Evaluate the cost/benefit of all possible pruning actions}
    \For{each action $A_i$ in $\mathcal{A}$}
        \State $\Delta \mathcal{L}_{\text{MSE}_i} \gets \text{eval}(\hat{\epsilon}_{\theta}^{\text{super}}, C_{\text{current}}, A_i)$
        \State $\Delta \#\text{Params}_i \gets \text{GetParamReduction}(C_{\text{current}}, A_i)$
    \EndFor

    \State \textit{// Execute the most efficient action (least performance drop per parameter removed)}
    \State $\hat{A} \gets \underset{A_i \in \mathcal{A}}{\arg\min} \frac{\Delta \mathcal{L}_{\text{MSE}_i}}{\Delta \#\text{Params}_i}$
    
    \State Update current configuration: $C_{\text{current}} \gets \text{ApplyAction}(C_{\text{current}}, \hat{A})$
\EndWhile

\State $C_{\text{optimal}} \gets C_{\text{current}}$

\Statex
\State $\rightarrow$ \textbf{Final fine-tuning of the searched architecture:}
\State $\hat{\epsilon}_{\theta}^{\text{optimal}} \gets \text{GetSubnet}(\hat{\epsilon}_{\theta}^{\text{super}}, C_{\text{optimal}})$ \Comment{Inherit weights from the super-net}
\State Fine-tune $\hat{\epsilon}_{\theta}^{\text{optimal}}$ with Knowledge Distillation. \Comment{As described in \cref{sec:KD feature align}}

\end{algorithmic}
\end{algorithm}

%% file: minipages/tiled_gemm.tex
\begin{figure}[ht]
\begin{minipage}[t]{0.475\textwidth}
\centering
\includegraphics[width=0.9\linewidth]{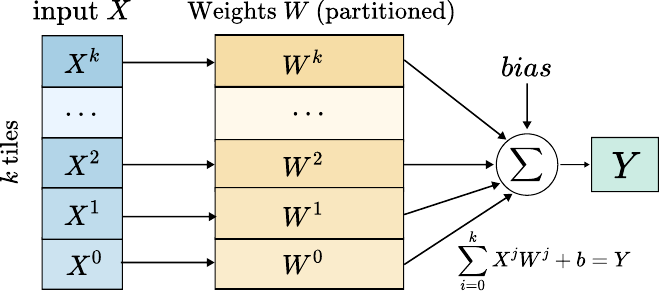}
\captionof{figure}{Illustration for tiled GEMM for a single token. The input $X$ and weights $W$ are both tiled into $k$ partitions along input feature.}
\label{fig:gemm tile}
\end{minipage}
\hfill
\begin{minipage}[t]{0.475\textwidth}
\centering
\includegraphics[width=0.85\linewidth]{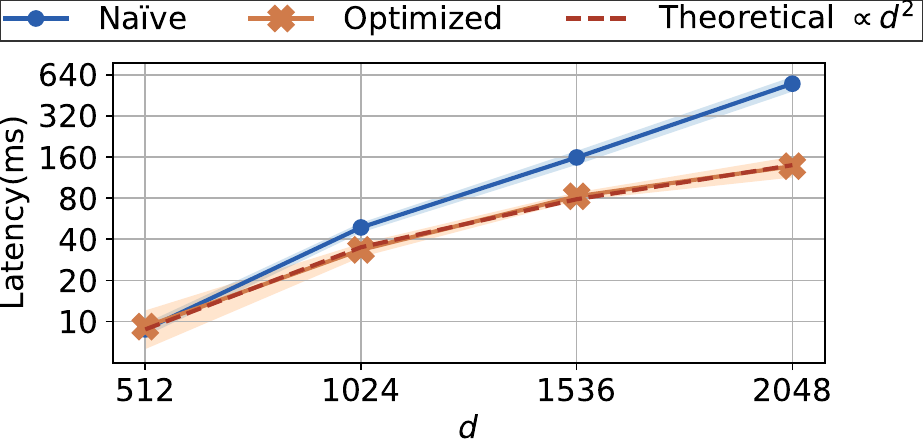}
\captionof{figure}{Latency benchmark for tiled GEMM in FFN. Note that the y axis is $log2$ scale.}
\label{fig:gemm tile bench}
\end{minipage}
\end{figure}



%% file: table/vae_variants.tex
\begin{minipage}[t]{0.5\textwidth}
    \centering
    \captionof{table}{Impact of VAE latent channels~$\left|z\right|$.}
    \resizebox{0.7\linewidth}{!}{
    \begin{tabular}{l|ccc}
    \toprule
       \textbf{VAE}  & $\left|z\right|$ & \textbf{PSNR} & \textbf{VBench}  \\
    \midrule 
    $4\times16\times16$ & 64 & 33.1 & 80.35 \\
    $4\times16\times16$ & 128 & 33.6 & 80.33 \\ \hline 
    $8\times32\times32$ & 128 & 30.6 & 79.73 \\
    $8\times32\times32$ & 256 & 30.8 & 79.80 \\ \hline 
    $8\times64\times64$ & 128 & 26.9 & 74.43 \\
    $8\times64\times64$ & 256 & 28.2 & 78.40 \\
    $8\times64\times64$ & 512 & 28.5 & 78.00 \\
    \bottomrule
    \end{tabular}
    }
    \label{tab:vae latent}
\end{minipage}
\hfill
\begin{minipage}[t]{0.5\textwidth}
    \centering
    \captionof{table}{Impact of VAE latent channel and patchify size for different compression strategy.}
    \resizebox{\linewidth}{!}{
    \begin{tabular}{l|ccccc}
    \toprule
       \textbf{VAE}  & \textbf{Patchify size} & $f$ & $\lvert z \rvert$ & \textbf{PSNR} & \textbf{VBench}  \\
    \midrule 
    $4\times8\times8$ & 2  & \multirow{2}{*}{1024} & 16& 33.2 & 80.32 \\
    $4\times16\times16$ & 1  & & 64& 33.1 & 80.35 \\ \hline 
    $4\times8\times8$ & 4  & \multirow{3}{*}{4096} & 16& 33.2 & 80.08 \\ 
    $4\times16\times16$ & 2  & & 64& 33.1 & 80.19 \\ 
    $4\times32\times32$ & 1  & & 256& 30.9 & 79.95 \\ 
    \bottomrule
    \end{tabular}
    }
    \label{tab:vae variant}
\end{minipage}

%% file: table/distill_steps.tex
\begin{table}[ht]
\centering
\caption{Ablation study on different inference steps.}
\resizebox{0.35\linewidth}{!}{
\begin{tabular}{l|ccc}
\toprule
\textbf{\#Steps} & \textbf{Quality} & \textbf{Semantic} & \textbf{Total} \\
\midrule
1 & 74.95 & 64.41 &72.84 \\
2 & 78.70 & 69.92 & 76.94 \\
4 & 83.81 & 72.89 & 81.63  \\
8 & 83.89 & 73.12 & 81.74  \\
\bottomrule
\end{tabular}
}
\label{tab:distill-step}
\end{table}

%% file: table/latency_breakdown.tex
\begin{table}[!ht]
    \centering
    \caption{Latency breakdown for on-device demo.}
    \resizebox{0.55\linewidth}{!}{
    \begin{tabular}{l|cccc}
        \toprule
        Module & Text Encoder & DiT & VAE decoding & I/O and Misc \\
        \midrule
        Latency~(ms) & 6  & 668\textsuperscript{\dag} & 230 & 320 \\
        \bottomrule
        \multicolumn{5}{l}{\textsuperscript{\dag}The latency for the DiT is corresponds to a single denoising step.}
    \end{tabular}
    }
    \label{tab:latency breakdown}
\end{table}

%% file: figure/xcode_benchmark.tex
\begin{figure}[ht!]
    \centering
    \includegraphics[width=0.9\linewidth]{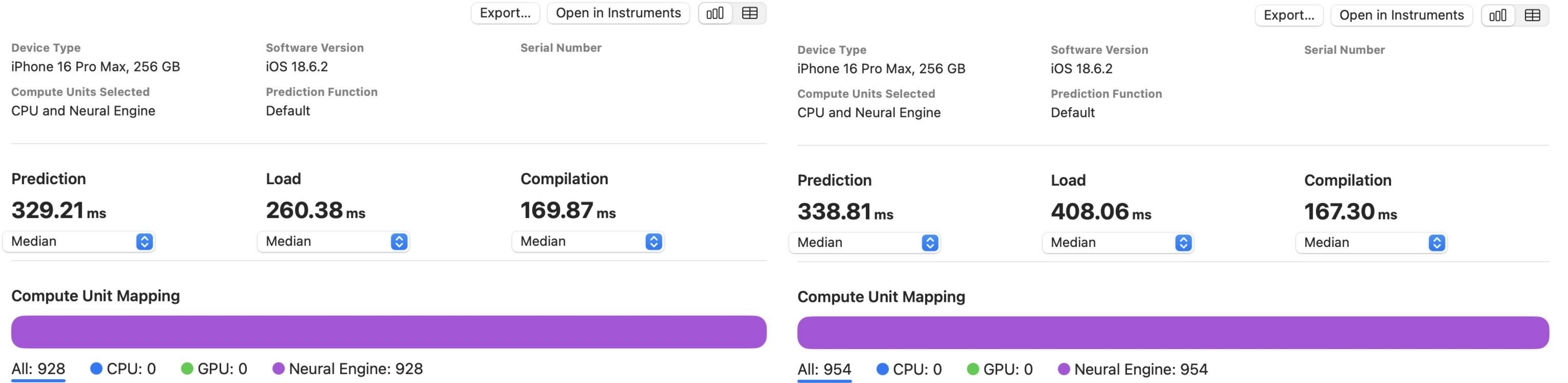}
    \label{fig:coreml}
    \caption{Latency Benchmark on Apple’s CoreML toolkits within Xcode.}
\end{figure}

%% file: figure/more_results.tex
\begin{figure}[ht!]
    \centering
    \includegraphics[width=\linewidth]{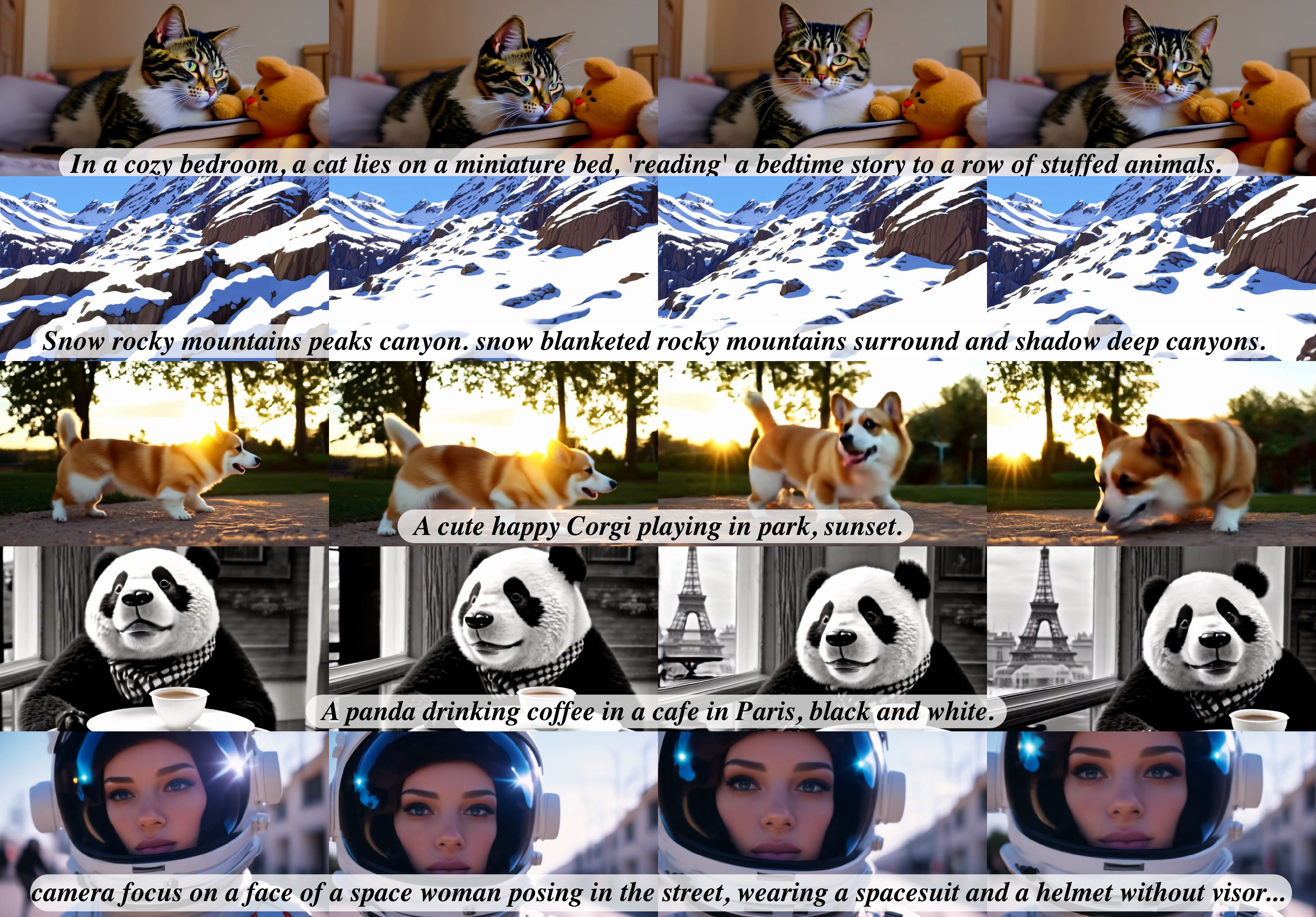}
    \caption{More quality results generated by our server model.}
    \label{fig:more server results}
\end{figure}

\begin{figure}
    \centering
    \includegraphics[width=0.9\linewidth]{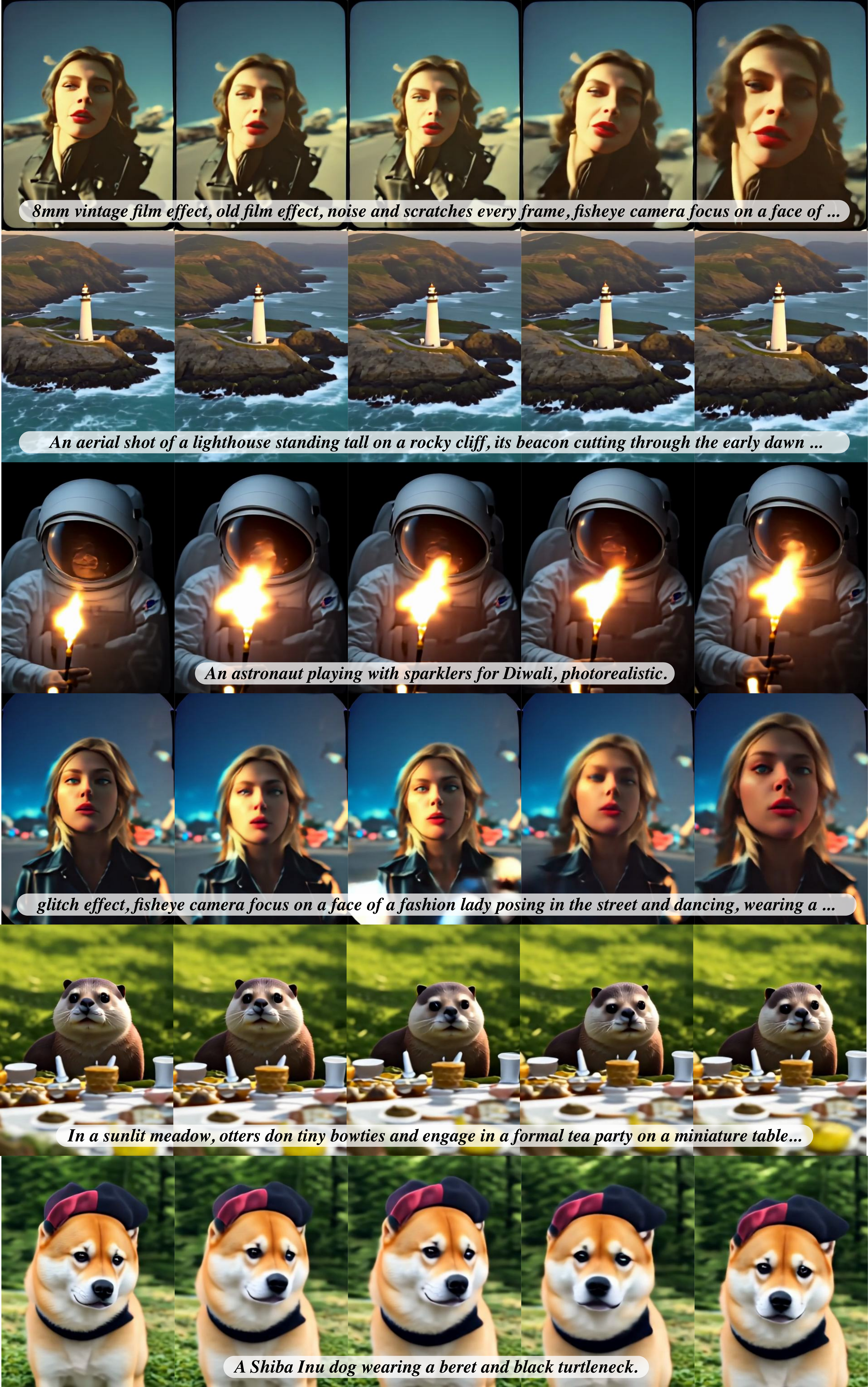}
    \caption{More quality results generated by our mobile model.}
    \label{fig:more mobile results}
\end{figure}